# Assessing the State of AI Policy

Joanna F. DeFranco, *Senior Member, IEEE* and Luke Biersmith

*Abstract*— The deployment of artificial intelligence (AI) applications has accelerated rapidly. AI enabled technologies are facing the public in many ways including infrastructure, consumer products and home applications. Because many of these technologies present risks either in the form of physical injury, or bias, potentially yielding unfair outcomes, policy makers must consider the need for oversight. Most policymakers, however, lack the technical knowledge to judge whether an emerging AI technology is safe, effective, and requires oversight, therefore policy makers must depend on expert opinion. But policymakers are better served when, in addition to expert opinion, they have some general understanding of existing guidelines and regulations. This work provides an overview [the landscape] of AI legislation and directives at the international, U.S. state, city and federal levels. It also reviews relevant business standards, and technical society initiatives. Then an overlap and gap analysis are performed resulting in a reference guide that includes recommendations and guidance for future policy making

*Index Terms*—AI Assurance, Artificial Intelligence, AI, AI Policy, AI Education.

## I. INTRODUCTION

POLICIES, regulations, standards, laws, trail advances in the associated, new technology. The delay is usually due to the time required to understand the safety and ethical consequences of the technology. That is, any policy framework is informed by science – and the science always occurs first. Before making claims resulting in regulations to address the claims, the science needs to occur first. When policy tries to predict science, it can result in confusion or worse. For example, soon after the introduction of the railway technology it was thought by some that the speed of a train (either riding or watching one) could cause mental unrest. In response, regulations were considered that required six-foot walls around the tracks. Historically this reaction to innovation is not unusual. The introduction of phones and electricity caused similar safety fears. But once the benefits and issues are understood, policy can be considered (e.g., Interstate Commerce Act in 1887 aimed to regulate and oversee the railroad industry [2]).

Recent attention to Chatbots highlight the need for regulation given the widespread use and acknowledgement that this type of technology is "unchecked and unregulated" for inaccuracies, bias, discrimination, and cause privacy violations [3]. Often the challenge in creating laws and regulations is the lack of basic understanding from the law makers. In addition, there is a divergence around the interpretation of ethical principles – which can turn into uncertainty when prioritizing solutions to addressing ethical principles [4]. The purposes of this work, then, is to serve as a blueprint and facilitate a legal response for responsible and informed policy that manages the risk in using AI.

In this paper, a review of previous and related work is presented followed by a review of six categories of AI assurance initiatives: state legislatures, city entities, federal entities, international groups of governance and policy, private sector business and technical societies. Then an overlap and gap analysis is conducted resulting in a reference guide that includes recommendations and guidance for future policy making.

## II. PREVIOUS AND RELATED WORK

Given the rapid acceleration of AI technologies over the past few years and their increasing integration with everyday life, the relationship between technology and laws, policies, and regulations has been a consistent subject of debate and study. There have been hefty fines for companies that violate data privacy laws (e.g., HIPAA, GDPR) but privacy violations are not the only risk for AI. There are major concerns such as safety, fairness, and ethics at stake. This section aims to provide a brief but substantive analysis of previous work related to this topic.

AI Ethics in the Public, Private, and non-government (NGO) Sectors [5], aims to map the ethical implications of AI by looking at ongoing work across three distinct sectors. The research prioritized principles, frameworks, and policy strategies aimed at the ethical use and implementation of AI technologies. By examining over one hundred documents from 25 different countries across a set time frame, the authors of this research were able to identify key differences in how AI technologies are being discussed in public, private, and NGO spheres. The paper aggregated ethics codes, principles, frameworks, guidelines, and policy strategies in order to understand how different organizations and sectors view AI ethics, their role in society, and what frameworks exist to manage them. The results of the research analyzed key differences by sector through the following topics: participatory engagement, engagement with law and regulation, consensus topics or areas of agreement, ethical breadth and depth, prominent sectoral differences, and finally omitted topics. The work concluded that public and NGO documents or frameworks were much more participatory in their creation and engagement with the law whereas the private sector was more concerned with privacy and customer related issues [5]. Overall, the work provided a strong analysis of frameworks and guidelines pertaining to AI ethics that exist across the public, private, and NGO sectors.

In another study that performed a systematic assessment of the AI policies of 25 countries it was discovered that a large number of countries do not have or are in the process of developing an AI policy. In addition, the policies analyzed had significant differences but overlapped in frequently discussing two ethical principles: *justice & fairness* and *nonmaleficence* [6]. Jobin et. al. (2019), discovered convergence of these principles including 3 others (transparency, responsibility, and privacy) – but recognized the differing interpretation of these principles.

In AI Certification: Advancing Ethical Practice by Reducing Information Asymmetries [7], focuses heavily on analyzing the methods available to organizations and sectors to ensure their various AI technologies are meeting their ethical standards. Referring to this as "certification" in the research, it may also be known as a form of "assurance." The review focuses on current AI certifications and programs both technical and non-technical in nature. For example, a section covers key certification concepts such as object of certification, means-end decoupling, and credence attributes [7]. However, this work also studies guidelines, policies and frameworks from regulatory bodies, including a white paper on artificial intelligence from the European Commission and an analysis from Queen's University on principles of AI implementation. The research is then able to categorize the various certification programs as belonging to four distinct areas including self-certification of AI systems, third-party certification of AI systems, third-party certifications of individuals, and third-party certification of organizations. A general case is made for addressing information asymmetries in AI, incentivizing correct behavior, and providing sound arguments for the need to integrate certification in technical areas and non-technical areas alike, such as public policy.

AI assurance research is prevalent given the large scope it includes as it refers to testing algorithms to be trustworthy, explainable, Safe, and ethical. Batarseh et. al. (2021) performed a systematic literature review on AI assurance research between the years of 1985 and 2021. The results provided insights, recommendations, and future direction for AI assurance. Specifically, they proposed five considerations when applying or defining and AI assurance method (1) data quality, (2) specificity, (3) addressing invisible issues, (4) automated assurance, and (5) the user. They also recommend multidisciplinary domain collaborations in the AI assurance research field to include other domains such as healthcare, education, and economics.

U.S. Artificial Intelligence Governance in the Obama-Trump Years by Adam Thierer [9] offers insight into how AI affects policymakers and what strategies were employed by two administrations that were objectively different in their approach to various policy topics. However, Thierer argues that both administrations shared a similar approach to AI technologies that functioned as a guide for policymakers and key stakeholders both in the public and private sectors [9]. Both administrations adopted what the author terms a "light touch" regulatory and industrial policy stance toward AI, focusing on key areas of concern for policy makers – safety and security. Both administrations looked for a growth first approach to policy rather than something that was highly restrictive or looked to limit the advancement of technology. The research also contends that both the Obama and Trump administrations differed from their China and EU counterparts, who have chosen a path of more hard law or broad regulatory frameworks covering AI and related technologies. The research provides an extensive background on how policy oversight of advanced technologies has progressed, including older regulatory frameworks from the Clinton administration in the mid-1990s that covered Information and Communication Technologies or ICT. Previous administration approaches had focused on hard-law applications including restrictions on entry, price controls, equipment regulations, and different quality-of-service requirements [9]. As time progressed and AI began accelerating during the Obama administration, the author reasons that a soft touch was being utilized with the intend to let AI bloom rather than preemptively restrict it. The Obama administration's approach to AI was reviewed by various policy groups and stakeholders, including a subcommittee on machine learning and AI (MLAI) as well as the National Science and Technology Council (NSTC) overseen by the White House Office of Science and Technology Policy (OSTP). The research makes the case that the Trump administration had mostly continued the policies of the Obama administration, promoting a pro-growth approach to AI. The hallmark of the Trump administration as it relates to AI is certainly Executive Order 13859 or "Maintaining American Leadership in AI". Similar to the Obama administration, this aimed to establish a more national approach to AI, pulling together both private and public resources. Thierer then analyzes difference between the U.S. approach under both administrations and that of the EU and China; most notably, the author found that the EU and China have developed a more robust legal framework for managing AI technologies, differing dramatically from any "soft touch" approach. Overall, the research offers policymakers an insight on how to approach AI technologies with examples from previous administration's lawmakers and key stakeholders.

Given that there are several different terms used throughout this paper that reference AI and related technologies, it is important to conclude this section with a brief explanation on two in particular: *assurance* and *policy*. In the context of this paper, "assurance" refers to the different technical and often complex methods used to test the reliability, safety, and efficiency of AI systems and methods. Generally, this includes stated goals for AI, including that it is explainable, trustworthy, and ethical for example [10]. Regarding the term "policy", this is an umbrella term used throughout the paper to simply highlight the different laws, standards, executive orders, or private sector guidelines that may guide a policymaker's understanding of AI and AI assurance. The aforementioned works represent only a sample of the work on the overlap between AI technologies and the policy space. The legal, regulatory, and technical frameworks that exist to govern these advanced technologies will be key resources for policymakers across a broad spectrum that includes public, private, and NGO organizations.

III. Methodology

The focus of this paper, as stated above, is to provide a reference for policymakers, stakeholders, and key individuals who wish to understand AI policies, AI assurance and its related concepts. Given that AI and AI assurance are part of a broad field of analysis and study, careful consideration was paid to the methodology behind gathering and organizing information to be included. This section explains both the "what" and the "how" that guided the organization of this paper. It became clear that first developing the structure, or the "what" in this case, would shape the "how" of knowing which sources and information to include. An echelon-based approach using grey literature (i.e., not indexed in conventional scholarly databases) was utilized, first looking at assurance practices at the city level and then gradually scaling up to include state, federal, international as well as industry standards and technical societies. This approach ensured a full picture of the AI assurance landscape, focusing on both the different levels of information as well as different areas, namely public and private sector.

Once the structure was developed, it was important to establish search criteria and methodology. Most searches were completed using open-source methods, namely Google as well as various journals and academic databases. Searches focused on broad terms, including "AI", "AI assurance", "assurance methods", "government AI assurance", "AI policies", and "business AI assurance". Sources utilized also focused on peer-reviewed journal and scholarly articles. This sourcing method yielded substantial results, particularly for the section on industry standards.

Once the various search results were complete, the next step was to identify inclusion and exclusion criteria. Inclusion criteria were straightforward once the structure of the article was complete, resulting in sources being placed in one of the five echelons identified above. Exclusion criteria was more subjective however, with particular attention being paid to any information that was deemed too technical or complex.

Given that the focus of this paper is geared towards policymakers and other non-technical stakeholders, reducing complexity remained the goal of the exclusion criteria. Information on AI and AI assurance practices that was considered more technical or academic in nature was not included. With the structure of the paper built out and the guidelines for sourcing information finalized, the "what" and the "how" were complete.

## IV. AI Assurance Initiatives

The large scope of AI and AI assurance issues need to be easily accessible and understandable. In this section we provide an overview of the current landscape of AI assurance initiatives that includes existing laws, directives, reports, and working groups across sectors would be beneficial to assist in policymaker understanding across several categories.

*A. U.S. Federal*

There has been significant activity by the U.S. government with respect to AI and AI assurance, and directives and initiatives have been launched across a broad range of agencies. A sampling of these appears in Table 1 (links to policies available https://docs.google.com/document/d/1G39rQhClospG7d294rqf90PPHF-2tKb5/edit ).

TABLE I
AI Assurance Taxonomy for Policymakers – U.S. Federal Level.

| Policy/Directive (sorted) | Source | Summary |
|---|---|---|
| **E.O. on Safe, Secure, & Trustworthy Artificial Intelligence (2023)** | Executive Office of the President | Establishes new standards for AI safety and security while focusing on privacy concerns, advancing equity and civil rights, and supporting a diverse and competitive workforce. |
| **E.O. 13859 (2019)** | Executive Office of the President | Establishes federal principles and strategies to strengthen the nation's capabilities in artificial intelligence (AI) to promote scientific discovery, economic competitiveness, and national security. |
| **NIST Report in Response to E.O. 13859 (2019)** | NIST | Calls for federal agencies, led by NIST, to bolster AI standards-related knowledge, leadership, and coordination among agencies that develop or use AI; promote focused research on the trustworthiness of AI systems; support and expand public-private partnerships; and |
| | | engage with international parties. |
| **Artificial Intelligence Standards Coordination Working Group (AISCWG)** | NIST Interagency Committee on Standards Policy (ICSP) | Facilitates the coordination of federal government agency activities related to the development and use of AI standards to the ICSP as appropriate. AISCWG activities also support NIST's Federal Coordinator role for AI standards. |
| **Public Law 116–283 Section 5301** | National Defense Authorization Act for Fiscal Year 2021 | Outlines the importance of artificial intelligence initiatives laid down for the Department and Commerce and by extension NIST. Prioritizes NIST as the federal lead for all AI and assurance standards. |
| **H.R. 7900** | National Defense Authorization Act of 2023 | Address certain AI needs such as testing, validation and education. |
| **U.S. DoD responsible AI Strategy and Implementation Pathway** | Department of Defense | A strategic approach to operationalize AI ethical principles. |
| **Artificial Intelligence Accountability Framework(2021)** | Government Accountability Office | Identifies key accountability practices – centered around the principles of governance, data, performance, and monitoring – to help federal agencies and others use AI responsibly. |
| **Principles of AI Ethics for the Intelligence Community (2020)** | Office of the Director of National Intelligence | Intended to guide personnel on whether and how to develop and use artificial intelligence, to include machine learning, in furtherance of the intelligence community's mission. |
| **AI: Overview, Recent Advances, and Considerations for the 118th Congress (2023)** | Congressional Research Service | Outlines current technical and policy considerations concerning AI and its related technologies and the federal involvement required for understanding what must come next. |

The broadest of these, E.O. 13859, is commonly known as the "Maintaining American Leadership in Artificial Intelligence" executive order. E.O. 13859 is one of the first directives aimed at outlining a national approach to AI. It recognizes the importance of AI as it relates to economic health, global competitiveness, and national security. The executive order lays out a coordinated federal strategy through the American AI Initiative, which is guided by five principles these are:

- the need for technological breakthroughs,
- the development of technical standards,
- training future generations,
- fostering public trust,
- and promoting an international environment that supports innovation and research.

This executive order also tasks NIST, via the Department of Commerce, to issue a plan for federal engagement in the development of technical standards and tools in support of trustworthy AI i.e., assurance practices. In response, NIST has already launched initiatives and developed a framework to address AI Risk Management (AI RMF) [11][12]. The AI RMF is intended to address risk by providing recommendations to govern, map, measure, and manage the risks of AI in the design, development, use, and evaluation of AI products, services, and systems – thus increasing AI trustworthiness. The AI framework integrates the ISO/IEC TS 5723:200 (en) standard – where the characteristics of trustworthiness (i.e., accountability, accuracy, authenticity, availability, controllability, integrity, privacy, quality, reliability, resilience, robustness, safety, security, transparency, and usability) are defined – allowing for terminology consistency, clarity, and understanding within the AI RMF [12].

Finally, the National AI Initiative Act of 2020 calls for a cross-cabinet level committee, the National Artificial Intelligence Advisory Committee (NAIAC). The NAIAC is tasked with advising the President and the National AI Initiative Office on topics related to the National AI Initiative. This Advisory Committee was launched in April 2022.

*B. U.S. State*

According to the US Chamber of Commerce, every state has AI legislation either enacted or pending, in 2021 (see Figure 1). These legislative efforts are important for systems builders and users alike.

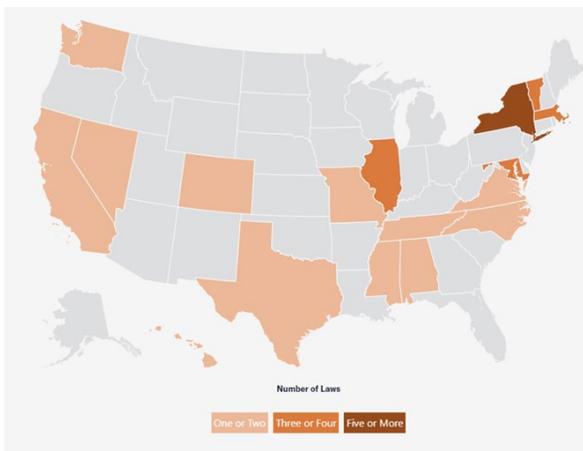

Interactive https://dev.uschamber.com/technology/state-by-state-

**Fig. 1.** State AI legislation activity in 2021 [13].

In 2022, AI bills or resolutions were introduced in seventeen states. The National Council of State Legislatures tracks state AI legislation. A sample appears in Table 2.

TABLE II
A sample of state legislative efforts to provide AI Assurance to the public [14].

| Legislation | State | Summary |
|---|---|---|
| Assembly Bill 2408 | California | Prohibits social media from using a design, feature, or affordance that the platform knew, or by the exercise of reasonable care should have known, causes a child user to become addicted to the platform (Pending) |
| Senate Bill 113 | Colorado | Creates a task force for consideration of facial recognition services, to recommend whether the scope of the task force should be expanded to include consideration of AI (Enacted). |
| Senate Bill 252 | Illinois | Creates the Innovation, Automation, and Structural Pilot Program to promote innovation through the combined application of tech, AI, automation, and distributive generation. (Pending) |
| Senate Bill 46 | Massachusetts | Requires covered entities whether they use automated decision systems. (Pending) |
| House Bill 4439 | Michigan | Requires review of computer system algorithms and logic formulas used by the unemployment security agency. (Pending) |
| Assembly Bill 168 | New Jersey | Requires the Commissioner of Labor and Workforce Development to conduct study and issue report on impact of artificial |

| | | intelligence on growth of state economy. (Pending) |
|---|---|---|
| **Senate Bill 5904** | New York | Prohibits motor vehicle insurers from discrimination on the basis of socioeconomic factors in determining algorithms used to construct actuarial tables, coverage terms, premiums and/or rates. (Pending) |
| **House Bill 1338** | Pennsylvania | Establishes the Automated Decision System Task Force (Pending) |

For example, Alabama's Senate Bill 78 is "aimed at keeping pace with innovation". S.B. 78 established the Alabama Council on Advanced Technology and Artificial Intelligence to "to review and advise the Governor, the Legislature, and other interested parties on the use and development of advanced technology and artificial intelligence in this state." This legislation passed with significant bipartisan support.

The Council has fourteen members appointed by the Governor, including the Secretaries of Commerce and Information Technology. All members or the council, excluding legislators, are required to have expertise in one of the following areas: artificial intelligence, workforce development, technology, ethics, privacy, or computer science. S.B. 78 also requires yearly reports to the governor on any recommendations for policy relating to artificial intelligence and advanced technologies. S.B. 78 is representative of AI and AI assurance initiatives being implemented at the state level.

*C. City*

Modern cities represent an ecosystem composed of residents, businesses, schools, and government. As such, cities need to establish a strategy to not only take advantage of the technological opportunities but also protect its citizens from danger. Therefore, a city has a responsibility to create a governance structure and strategy for the development and application of AI in its city's ecosystem – in particular how the city can most effectively, efficiently, and ethically perform AI risk assessment, use and development of AI systems used in their city.

Some cities have not published their own strategy which may indicate defaulting to their state or countries strategic plan/effort towards AI. For example, London, usually listed at in the top five wealthiest cities, uses the UK's guidance [15] for AI and the city of Moscow uses Russia's AI strategy [16]. The challenge in creating a response to innovative technology is understanding it. In other words, although they understand there is risk – they do not understand the technology causing the risk. Lack of understanding slows the creation of laws and regulations as the risks and issues need to be understood before consensus is achieved, and unfortunately, some law makers lack in understanding basic AI concepts [17].

Some cities have developed their own AI strategies. An analysis was performed on AI strategies of three cities that represent major parts of the world (New York City/US, Hong Kong/Asia, and Paris/Europe). The results showed that NYC created a 5-part AI strategy [18], Paris established a 4-part action plan [19], and Hong Kong developed AI guidance consisting of a 4-part practice guide [20]. A qualitative analysis was performed on those results and determined the emerging five overlapping themes among the three AI strategies (Table 3).

TABLE III
ANALYSIS OF CITY AI STRATEGY

| Initiative overlaps | Highlights |
|---|---|
| **NYC: Improve city data infrastructure** <br><br> **HK: Manage quality of the data used to train AI models should be managed** | NYC: Goals to build on data collection, acquisition, procurement, standards, cleaning, sharing, ownership access, usage agreements, warehousing, and analytics, digital rights concerns, such as privacy and security. For example, data should be collected in a "machine readable" manor and shared with others to increase the range of use. <br><br> HK: AI data prep process: improve quality (e.g., annotation, labelling, cleaning, enrichment, and aggregation). |
| **NYC: City governance and policy around AI** <br><br> **HK: Governance Structure** <br><br> **Paris: Democratic, multi-scalar, dialogue for international AI** | NYC: Only initial foundation of AI governance has been developed. Acknowledge what is unknown. <br><br> HK: Develop a committee to oversee AI development, use and termination. |
| governance | Paris: Facilitate dialogue toward concrete AI principles for AI governance based on a vision of human rights. |
| **NYC: Accountability**  **Paris: National measures for corporate accountability in AI-based services**  **HK: Accountability** | NYC: Be responsible for outputs, decisions or impacts resulting from the use of AI.  Paris: Accountability of transnational corporations in AI-based services (e.g., source code disclosure to appropriate authorities to protect human rights and market abuse).  HK: Set up specific internal policies and procedures on how to design, develop and use AI ethically, including an institutionalized decision-making process with escalation criteria |
| **NYC: Partnerships with external organizations**  **Paris: Collaboration**  **HK: Communication** | NYC: Agencies and external stakeholders desire stronger ways to collaborate.  Paris: A global database that tracks and monitors AI legislation for human rights and development implications, facilitating contextual policy making.  HK: Regularly communicate AI strategy, policies, and procedures to all relevant personnel, including internal staff at all levels and, where appropriate, external stakeholders such as business partners. |
| **NYC: (1) Business education (2) AI applications within the city**  **Paris: Retain domestic AI Talent**  **HK: Awareness raising** | NYC: (1) Leverage and coordinate efforts to develop a local AI workforce. (2) help agencies ID, assess, and realize use of AI in the city  Paris: Create incentives to retain domestic AI talent in the Global South and build local research and development capabilities.  HK: (1) hiring internal and external experts with relevant technical skills, experience, and expertise to develop and use the AI system. (2) training on compliance laws/regulations/policies and human reviewers in charge of overseeing the decision making of AI systems |

Between the three reports, there were overlapping themes in the AI strategy. For example, all three cities recognize the benefits of AI and had safety concerns with areas of emphasis such as ethical use, impacts, and risks of AI. The reports clearly communicate there could be danger or opportunity missed if risks were not assessed and development strategies were not well thought out and governed. Aggregating the overlapping themes resulted in six core considerations for a city's AI strategy shown in Table 4. The sixth core consideration came from differentiating the educational needs for the public/business as well as training to create a complete AI workforce.

TABLE IV
SIX CORE CONSIDERATIONS FOR A CITY'S AI STRATEGY

| | Consideration | Definition |
|---|---|---|
| 1 | Improve data quality and infrastructure | Build on data collection and quality |
| 2 | Develop a collaborative governance (domestic and global) on AI use | Create a dialogue toward concrete governance. Only initial foundation of AI governance has been developed. Acknowledge what is unknown |
| 3 | Develop accountability guidelines | Create internal policies (e.g., code discloser) and procedures on how to design, develop and use AI ethically (protect human rights and market abuse), including an institutionalized decision-making process with escalation criteria |
| 4 | Improve Communication | Develop stronger ways to regularly collaborate among agencies and external stakeholders |
| | | (e.g., global database to track and monitor AI legislation for human rights and development implications, facilitating contextual AI strategy and policy making |
| 5 | Improve awareness and develop education | Train on compliance laws/regulations/policies |
| 6 | Create an AI workforce | -Leverage, retain and coordinate efforts to develop AI Systems<br>-hire internal and external experts with relavant technical skills, experience, and expertise to use and REVIEW the AI systems.<br>-help agencies ID, assess, and realize use of AI in the city (across all sectors) |

*D. International*

Assessing international standards for AI and AI assurance is more difficult, given that civil liberties, privacy, data protection, and security norms and laws are different in each country and differ from that in the U.S. But there are consistent themes across the landscape of international AI assurance initiatives. International activity largely occurs in Europe and the Asia-Pacific sub-regions, where there are significant economic drivers for AI policy and innovation (see Table 5 - links to policies available https://shorturl.at/cgjDL).

TABLE V
AI ASSURANCE TAXONOMY FOR POLICYMAKERS – SAMPLING OF INTERNATIONAL EFFORTS. ADAPTED FROM [21]

| Policy/Directive (sorted) | Source | Summary |
|---|---|---|
| **Committee on AI/Guidelines on AI and Data Protection** | Council of Europe (COE) | Establishes a committee for the ethical use of AI and AI-related technologies that covers the European sub-region. COE's report seeks to provide a set of baseline measures that governments, AI developers, manufacturers and service providers should follow to safeguard humans. |
| **Policy and Investment Recommendations for Trustworthy AI** | EU High-Level Expert Group on AI (AI HLEG) | Part of the European Commission's broader AI strategy, the report describes 33 recommendations that can guide trustworthy AI towards sustainability, growth and competitiveness, and inclusion while empowering and protecting human beings. |
| **Intergovernmental Meeting of Experts (category II) related to a Draft Recommendation on the Ethics of Artificial Intelligence** | United Nations Educational, Scientific, Cultural Organization (UNESCO) | Addresses the implications of AI on UNESCO's central domains while adding a multidisciplinary, universal, and holistic approach to the development of AI in the service of humanity, sustainable development and peace. |
| **Principles for Responsible Stewardship of Trustworthy AI** | Group of 20 (G20) | Focuses on inclusive growth, sustainable development, and well-being. Recommend beneficial outcomes for people and the planet by augmenting human capabilities and enhancing creativity. |
| **Recommendations of** | OECD | Promotes the use |
| the Council on Artificial Intelligence | | of AI that is innovative and trustworthy and that respects human rights and democratic values. |
| **Responsible AI Working Group** | Global Partnership on AI (GPAI) | Fosters and contributes to the responsible development, use and governance of human-centered AI systems, in congruence with the UN Sustainable Development goals. Allows for the interfacing with other GPAI working groups and external governments. |

For example, the European Union and its sub-organizations have been a global leader in the research and application of AI assurance standards. The AI landscape includes laws, directives and reports, e.g.: *Coordinated Plan on AI 2021 Review, EU Cybersecurity Strategy, Digital Services Act, Digital Markets Act, Data Governance Act.* Although these contributions to the field are wide ranging, policymakers should focus on the Organization for Economic Cooperation and Development's AI Policy Observatory, specifically the *Recommendations on AI and Principles* report. Most importantly, this tool from OECD is an excellent example of engagement on the international level.

OECD AI Policy Observatory comprises over sixty nations and it maintains a repository of nearly 700 AI policy initiatives from said member nations; a useful resource for policymakers looking to obtain an international perspective on these topics. The OECD body lays out a set of value-based principles, including a focus on inclusive growth and sustainable development, human-centered values and fairness, transparency, safety, and accountability. This collection of AI assurance artifacts also incorporates specific recommendations for policymakers that cover a host of areas including the need to foster research and development and produce digital ecosystems for AI. Of all the international frameworks, laws, and standards included in the taxonomy, OECD's AI Policy Observatory may be the most far-reaching and detail rich of all and a must read for policymakers.

Advances in the Asia-Pacific region include development of the following: AI Ethics Framework, Standardization/Technology Framework (Australia), Beijing AI Principles, Next Generation AI Development Plan (China); Council on Ethical Use of AI and Data, Model AI Governance (Singapore); and Personal Data Protection Act, Ethical Guidelines for AI (Thailand).

*E. Industry Guidelines*

Many of the largest companies in the world, including Microsoft, IBM and Intel have succeeded at rapidly broadening our understanding of AI technology and have created various principles, reports and guidelines that highlight the need to make AI safe and reliable. A sample of these appears in Table 6 (links to policies available https://shorturl.at/cgjDL).

TABLE VI

SELECTED AI ASSURANCE GUIDELINES FROM BUSINESS. ADAPTED FROM [21]

| Initiative (sorted) | Company | Summary |
|---|---|---|
| **AI Ethics Guidelines** | Sony Group | Sets forth guidelines that must be followed by all officers and employees of Sony when utilizing AI and/or conducted AI-related research and development (R&D). |
| **Artificial Intelligence Principles** | Telefonica | Acknowledges the use of AI and big data to transform business. Focuses on fairness, transparency, privacy, and security as fundamentals of AI technology. |
| **Guidelines for Artificial Intelligence** | Deutsche Telekom | Defines nine self-binding guidelines that describes how Deutsche Telekom should use AI and how they intend to develop AI-based products and services in the future. |
| **Guidelines for Developers of** | Microsoft | Relates to the development and |
| Conversational AI | | use of "bots," Microsoft's guidelines aim to promote trust and efficiency while safeguarding privacy, data, and security. |
| Guiding Principles for Artificial Intelligence | SAP | Guiding principles to help steer the development and deployment of SAP's AI software to benefit people. Expands the conversation to include customers, employees, governments, and civil societies. |
| Guiding Principles for Ethical AI | Unity Technologies | Six guiding principles meant to serve as a blueprint for developers, business community, and vendors utilizing AI technologies. Focused on mitigating bias of AI systems. These principles are to be unbiased, accountable, fair, responsible, honest and trustworthy. |
| Principles on Artificial Intelligence | Google | Addresses Google's commitment to develop technology responsibly and establish specific application areas they will not purse through six principles described shortly. |
| Recommendations for U.S. National Strategy on AI | Intel | Direct response to E.O. 13859 outlining key challenges and recommendations needed to establish safe, responsible, and competitive AI systems and structures. |
| Everyday Ethics for Artificial Intelligence | IBM | IBM's effort to define, design, and development applicable ethics relating to AI for the technical community. |
| Report on Lessons in Practical AI Ethics | Digital Catapult | Creates an applied and practical methodology for machine learning ethics, designed for business and individuals wanting to adopt an ethical and responsible approach to machine learning development. |

Google is one of the largest companies in the world and a leader in research and development of AI technology. Google has outlined the following seven objectives for its AI:

1. Be socially beneficial
2. Avoid creating or reinforcing unfair bias
3. Be built and tested for safety
4. Be accountable to people
5. Incorporate privacy design principles
6. Uphold strict standards of scientific excellence
7. Be made available for uses that accord with these principles

These principles are very consistent with those framed in many state and Federal AI assurance initiatives.

*F. Professional and Technical Societies*

Numerous professional and technical organization, such as the IEEE, have produced AI and AI assurance guidelines. As sample appears in Table 7 (links to policies available https://shorturl.at/cgjDL).

TABLE VII
SELECTED AI ASSURANCE INITIATIVES FROM PROFESSIONAL AND TECHNICAL SOCIETIES. ADAPTED FROM [21]

| Initiative | Source | Summary |
|---|---|---|
| Harnessing AI: Recommendations for Policymakers | Information Technology Industry Council (ITIC) | Promotes the responsible development and use of AI technologies that will benefit people, society, and the economy. |
| Artificial Intelligence and Machine Learning Policy Report | Internet Society (ISOC) | Introduces AI basics for policymakers and other stakeholders. Highlights key considerations and challenges while issuing recommendations and high-level principles to follow. |
| Asilomar AI Principles | Future of Life Institute | Outlines key challenges and risks of AI technologies while promoting thirteen ethical principles and values critical to the use of AI. |
| A data brief on China's AI Strategy | China's Artificial Intelligence Industry Alliance (AIIA) | Makes clear the rights and obligations at each stage in artificial intelligence research and development. |
| Artificial Intelligence Research and Development Strategic Plan | Software and Information Industry Association (SIIA) | Draft of guidelines that supports a global definition for AI developed with public and private sector stakeholders to assist policymakers. |
| Business Ethics and Artificial Intelligence Report | Institute of Business Ethics (IBE) | Examines the ethical challenges for business in developing artificial technologies and suggests measures which can be adopted to minimize the risk of ethical lapses due to improper use of AI technologies. |
| Principles for Accountability Algorithms | Fairness, Accountability, and Transparency in Machine Learning (FAT ML) | Focuses on the human aspect in the development of algorithms that promote responsibility, explainability, accuracy, auditability, and fairness. |
| Report on Algorithmic Transparency and Accountability | U.S. ASSOCIATION FOR COMPUTING MACHINERY (USACM) | Support the benefits of algorithmic decision making while addressing concerns over harm, bias, and privacy. |
| *Various* | INSTITUTE OF ELECTRIC AND ELECTRONICS ENGINEERS (IEEE) | See forgoing discussion. |

To focus on how these technical societies are addressing these broad issues, consider, for example the IEEE, which is "the world's largest technical professional organization for the advancement of technology." IEEE has many standard and guidance initiatives the landscape of AI assurance, consider this sampling.

- IEEE P7001 – Transparency Of Autonomous Systems,
- IEEE P7003 – Algorithmic Bias Considerations,
- IEEE P7006 – Personal Data AI Agent Working

Group
- IEEE P7007 – Ontological Standard for Ethically driven Robotics and Automation Systems,
- IEEE P7008 – Standard for Ethically Driven Nudging for Robotic, Intelligent and Autonomous Systems ("Nudges" are overt or hidden suggestions or manipulations designed to influence the behavior or emotions of a user),
- IEEE P7009 – Standard for Fail-safe Design of Autonomous and Semi-autonomous Systems
- IEEE Std 7010 – Recommended Practice for Assessing the Impact of Autonomous and Intelligent Systems on Human Well-Being, and
- IEEE P7014 – Standard for Ethical considerations in Emulated Empathy in Autonomous and Intelligent Systems.

While most of these resources are highly technical, policymakers can rely on these and similar technical guidance documents and initiatives for term definition. Artifacts such as those produced by professional and technical societies should also form the technical foundation of any policymaking initiatives and be consistent any legislative and legal requirements.

In addition, these technical foundations likely integrate scholarly research on relevant topics. The collaborators on those documents perform their own research and understand critical issues such as the improper use of AI technologies. An example of scholarly research on a topic discussed in one of the Table VII reports, is about an online experiment where it was determined that algorithmic decision making (ADM), although can produce high-quality outputs, may be perceived as illegitimate when used by policy makers and as such should be used as only an assistant to human decision making [22].

V. Overlap, Gap Analysis and Roadmap for Future Guidance

To develop a roadmap for responsible AI, we qualitatively analyzed the six category tables for emerging themes. This analysis resulted in eleven themes outlined in table 8. The categories have some expected overlap across the initiatives described.

There are also some appropriate non-overlapping categories, for example, one of the Federal governments and International main concerns is to establish strategies to stay competitive, thus, it is appropriate that the other categories drill down into initiative areas that support the overarching effort to stay competitive such as developing technology that benefits society as a whole – augmenting human capability, society, and the economy. The city category has some non-overlapping categories involving focus on law and compliance training as well as developing a workforce that can evaluate, apply, and effectively use the available AI technology.

TABLE VIII
CATEGORY ANALYSIS

| Responsibility Initiative | Fed | State | City | Intnl. | Indus. | Prof Soc. |
|---|---|---|---|---|---|---|
| Strengthen Res | X | | | X | | |
| Governance | X | | X | | | |
| Coord/Comm | X | | | X | | |
| Assurance | X | | | X | X | |
| Ethics | X | | | X | X | X |
| Deve/use Guidance | X | X | X | | | |
| Sec./Privacy | | X | | X | X | X |
| Fairness, non-Biased | | X | | | X | X |
| Education | | | X | | | |
| Workforce | | | X | | | |
| Benefit as a whole | | | | X | X | X |

VI. AI Training for Executives and Legislators

Policymakers have a wealth of resources available to consult in efforts to reduce the potential for AI technology risks for the public. Given that policymakers at all echelons are entrusted with the power to shape the course of AI technology in the future, it is critical that technical and non-technical individuals alike can adequately understand such complex topics as AI and AI assurance. In this regard, education is key and in particular policy makers need to understand that public facing AI, should be explainable, safe, secure, trustworthy, ethical, and fair [23]. It is also important to place a focus on the competencies and skills required of the work force in the AI Sector. The United Nations Educational, Scientific and Cultural Organization (UNESCO), along with urging governments to implement a global AI Ethical Framework, has as a "Readiness Assessment" tool available for all to determine the skillset necessary to keep in line with AI regulations [24].

There are many online opportunities for AI education. For policy makers it is important to differentiate between technical and nontechnical courses and certifications. For example, there are online course offerings in artificial intelligence and machine learning basics for non-technical professionals (e.g., IEEE, Stanford)

VII. Policy Considerations for AI and AI Assurance

Despite considerable progress being made on AI assurance at all levels, interpreting, reconciling and applying these can be overwhelming as the policy makers face the following risks:

1. harmonizing the requirements of existing legislation, professional standards, and other mandates and guidance on artificial intelligence,
2. keeping up with the pace of change,
3. and technical knowledge gaps limiting AI technology understanding.

Research in AI is very robust with new theory and applications being developed continuously. Particular policy attention should be paid towards the types of AI deemed "frontier AI" models and technologies, those that are highly complex and promising yet still pose a considerable danger due to their unknown capabilities [25]. These can deploy to public facing applications, including critical ones, very quickly, sometimes without adequate verification and validation. This phenomenon can create a "regulatory gap" causing legislation, guidance and policy to lag behind technological developments. Despite their benefits, current technologies, such as self-driving cars, social media algorithms, drones, and other advancements may require oversight and regulation, which will represent a continuous challenge for policymakers.

## VIII. Discussion and Conclusions

The need for AI expertise in policy makers is real. When analyzing the current landscape, policymakers need to be educated on the key tenets of AI assurance in order to facilitate United States AI competitiveness while also safeguarding the safety and dignity of the citizens. Without individuals who can communicate the practical benefits and risks to the average citizen, these technologies may seem "scary." Therefore, policymakers need to be educated on the key tenets of AI assurance in order to facilitate United States AI competitiveness while also safeguarding the safety and dignity of the citizens. Legislators and policymakers can rely on experts, but they should have some minimal understanding of AI.

We will emphasize that Individuals with non-technical backgrounds can make a significant impact on technical fields, given that they understand their proper role and function. Following are recommendations to facilitate makings this impact:

- Emphasize effective inter-agency cooperation
- Engage the expertise of professional engineers (PE) to take some of the responsibility.
- Select committees in both the House and Senate designed *specifically* for the research, assessment, and development of current AI technologies and assurance standards
  - Committees should balance between both technical and non-technical legislators with assistance from outside sources
- Presidential Council on Artificial Intelligence Technologies would place the topic at the forefront and should act on developing the American plan for remaining globally competitive in AI while ensuring safety and reliability
  - For example, use Alabama S.B. 78 as a frame of reference for an effective, bipartisan way forward
- Continue to utilize budgetary tools to fund scientific research and development, particularly for younger populations
- Prioritizing the development of STEM fields will have an enormous impact on closing technical "knowledge gaps"

This list clearly can be expanded to implement responsible AI by helping to inform legislators, policy makers and other responsible government officials at all levels.

## Acknowledgment
Thanks to Phil Laplante for various discussions, recommendations and contributions that led to the expansion of the conference paper to the present work.

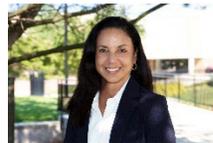
**Joanna F. DeFranco** (Senior Member, IEEE) earned her Ph.D. in computer and information science from New Jersey Institute of Technology, M.S. in computer engineering from Villanova University, and a B.S. in Electrical Engineering and Math from Penn State University.

She is an Associate Professor of Software Engineering at The Pennsylvania State University as well as the Associate Director of the Doctor of Engineering degree program. She also works as a researcher at the National Institute of Standards and Technology. She has worked as an Electronics Engineer for the Navy as well as a Software Engineer at Motorola. Her research interests include Internet of things, Distributed Systems, Critical Systems, and Teamwork.

Dr. DeFranco is a senior member of the IEEE. She is Associate Editor in Chief of IEEE Computer magazine as well as the IoT column and Fundamental Computing editor. She is on the editorial board of Security and Privacy Magazine and currently serving on the administrative committee for the IEEE Reliability Society.

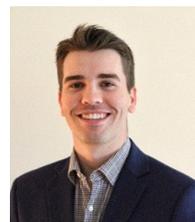
**Luke E. Biersmith** earned his Master of Public Policy with a focus on International Security and Economic Policy from the University of Maryland, and his B.A. in History from Penn State University. He is currently pursuing a Doctor of Law and Policy at Liberty University.

      He is an Assistant Vice President with Bank of America in Charlotte, NC working in Global Operations. He also serves as a military intelligence officer with the U.S. Army Reserve. His research interests include economic policy, foreign policy, and military history.